\documentclass{article}

\PassOptionsToPackage{numbers, compress}{natbib}
\usepackage[preprint]{neurips_2026}

\usepackage[utf8]{inputenc}
\usepackage[T1]{fontenc}
\usepackage{hyperref}
\usepackage{url}
\usepackage{booktabs}
\usepackage{amsfonts}
\usepackage{amsmath}
\usepackage{amssymb}
\usepackage{nicefrac}
\usepackage{microtype}
\usepackage{xcolor}
\usepackage{graphicx}
\usepackage{multirow}

\newcommand{\resolveatcost}{\texttt{resolve@cost}}
\newcommand{\ci}[2]{[#1, #2]}

\title{What Context Does a Coding Agent Actually Need to \emph{Act}?}

\author{%
  Brian Sam-Bodden\thanks{Code and data: \url{https://github.com/integrallis/act-context}} \\
  Integrallis Software \\
  \texttt{bsbodden@integrallis.com} \\
}

\begin{document}
\maketitle

\begin{abstract}
A modern coding agent can hold an entire repository in its context window. Most of its reading
is wasted --- and the interesting question is not how much context an agent can use, but what it
actually \emph{needs}. We study that question at the moment it matters most: when the agent must
\emph{edit} code. Separating \emph{finding} the work site from \emph{acting} on it, we hold
localization fixed with an oracle, vary only how the code is represented, and score context
against real issue resolution on SWE-bench Verified. The answer is starkly minimal. The signal
lives in the code being edited itself: natural-language summaries of it answer almost none of the
behavioral questions that the source answers ($4/45$ vs.\ $27/45$, held-out repositories,
independent judge), and the gap belongs to the representation, not the summarizer --- a frontier
model's summaries score exactly as poorly as a 3B model's. The surrounding context hardly matters
either: across every multi-file instance in Verified, under a protocol frozen before any data,
rendering a file's remainder as UML skeletons and signatures resolves no more issues than deleting
that remainder outright ($N{=}70$, exact McNemar $p{=}0.75$). That was our registered hypothesis,
and it failed. Compressed context, meanwhile, matches whole files at a third of the tokens: a
resolved issue costs $19$K context tokens, not $94$K. The instrument also yielded a finding the
field should keep: temperature-0 API inference flips ${\sim}9\%$ of per-instance outcomes between
byte-identical runs. That is a noise floor under every small effect reported on this benchmark,
including ours. We release the instrument --- gold-validated environments, per-instance proof that
every reference edit is expressible from every arm's context, deterministic patch construction,
and pre-registered hypotheses whose nulls we publish.
\end{abstract}

\begin{figure}[t]
  \centering
  \includegraphics[width=0.92\linewidth]{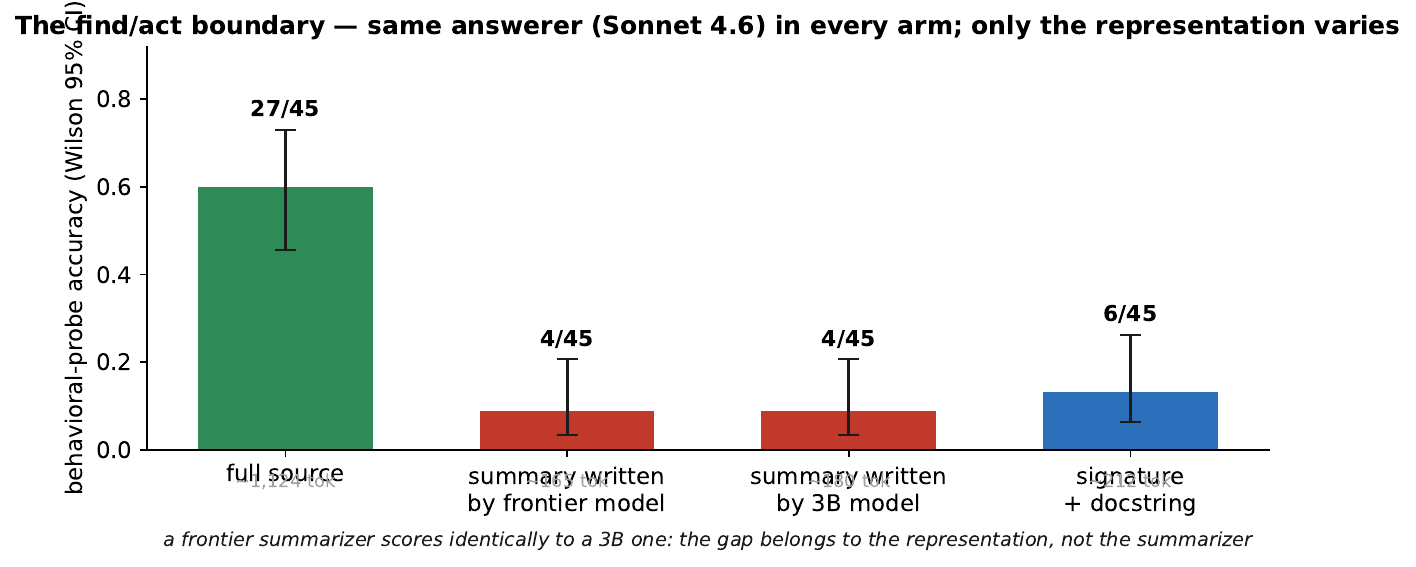}
  \caption{\textbf{The find/act boundary.} Behavioral-probe accuracy by context representation
  ($n{=}45$: 15 held-out classes from seaborn/pylint/pytest $\times$ 3 probes minted from each
  class's \emph{test file}; the \emph{same} model --- Claude Sonnet 4.6 --- reads the context and
  answers in every arm, so only the representation varies; answers graded by an independent-family
  judge, $\kappa{=}0.84$ against a second judge). The frontier/3B labels indicate the \emph{summarizer}
  that wrote the representation, not the answerer. Source answers $60\%$; natural-language
  summaries answer ${\sim}9\%$ --- identically for a frontier and a 3B summarizer --- and no
  summary succeeds where source fails (McNemar 23-vs-0, $p{<}10^{-4}$). Mean representation size in
  \texttt{cl100k\_base} tokens below each bar.}
  \label{fig:teaser}
\end{figure}

\section{Introduction}
\label{sec:intro}

When a software engineer fixes a bug, she doesn't re-read the repository. She finds the few
functions that matter, reads them closely, and lets the rest of the codebase fall into a vague map
of what exists where. Coding agents are built as if the opposite were true: context windows now
hold entire repositories, and the natural instinct is to fill them. That instinct has a
price --- tokens, latency, dollars --- and surprisingly thin evidence behind it. Frontier agents
already resolve a large fraction of SWE-bench Verified \citep{swebench}, yet ContextBench
\citep{contextbench} finds that sophisticated retrieval scaffolds often barely beat trivial
baselines, with a wide gap between context \emph{explored} and context \emph{used}. So the
question worth asking is not whether we can fit the code, but: \textbf{for a given code task, what
is the minimal context an agent actually needs --- and in what representation?}

We separate two sub-questions that are often conflated:
\begin{itemize}
  \item \textbf{Find} --- locating where the work happens;
  \item \textbf{Act} --- given the location, knowing enough to edit correctly.
\end{itemize}

To measure \emph{act}-context cost without rewarding ``include more,'' we hold localization fixed
with an \emph{oracle} (the same gold files for every arm) and vary \emph{only the representation}
(\S\ref{sec:method}). This yields a pre-specified prediction we can pass or fail: \emph{if program
structure is what an agent needs to act, a structured representation tier (UML-skeleton /
method-only) should resolve more, at equal localization, than a flat binary keep/drop}
\citep{swezze}. Section~\ref{sec:resolve} tests that prediction on every multi-file instance in
SWE-bench Verified, under a protocol frozen and committed before any data. \textbf{The prediction
fails}, and we report the null.

\paragraph{Contributions.}
(1)~An \textbf{audited \resolveatcost{} instrument}: oracle-localized representation arms
with per-instance \emph{expressibility} verification (every gold edit provably expressible from
every arm's context), per-instance gold-patch environment validation, deterministic patch
construction, and pre-registered hypotheses --- every number traceable to a raw artifact.
(2)~A \textbf{powered null}: structured representation tiers carry no resolve signal over binary
keep/drop at oracle localization, single-shot ($N{=}70$, $p{=}0.75$), while compressed context
matches whole files at $3$--$3.7\times$ fewer tokens.
(3)~A \textbf{de-circularized find/act boundary}: with test-minted probes, held-out repositories,
and an independent judge, source carries roughly $7\times$ the actionable signal of
natural-language summaries, and a frontier summarizer cannot close the gap (the first run of the
summarizer-quality control).
(4)~A \textbf{methodological caution} measured in passing: run-to-run non-determinism at $T{=}0$
flips ${\sim}9\%$ of SWE-bench per-instance outcomes between byte-identical runs --- bounding the
interpretation of all small arm gaps, ours and others'.

\paragraph{Rigor and scope.} Every number in this paper traces to a raw artifact. Protocols,
hypotheses, and instance samples were frozen and committed before data collection; every
instance's environment is validated with its reference patch before any model call; every
reference edit is verified expressible from every arm's context; figures regenerate from raw
result files. The \resolveatcost{} and probe experiments run on SWE-bench Verified and held-out open-source
repositories; superseded or exploratory measurements are quarantined in Appendix~\ref{app:explore}
and carry no claims. We
label every $n$, report a Wilson 95\% CI for every proportion, and state for each negative which
way its biases cut.

\section{Related work}
\label{sec:related}

\textbf{Sufficiency-based context for code.} SWEzze \citep{swezze} reduces issue contexts to a
minimal sufficient subsequence (binary keep/drop of code segments), validated on resolution;
Joren et al.
\citep{sufficientcontext} formalize ``sufficient context'' for RAG. We build on these; our
act-view probes ask \emph{which alternative rendering} suffices, and our \resolveatcost{}
experiment answers whether \emph{tiers} beat their \emph{keep/drop} (they do not, $N{=}70$,
\S\ref{sec:resolve}).
\textbf{Context pruning/compression.} SWE-Pruner \citep{swepruner}, LLMLingua \citep{llmlingua},
and related systems prune tokens or lines; we select among \emph{program-structure
representations} and measure \emph{resolution}, not relevance.
\textbf{Localization and retrieval.} Agentless \citep{agentless}, LocAgent \citep{locagent},
RepoGraph \citep{repograph}, and LingmaAgent \citep{repounderstander} address \emph{find}; we
control localization \emph{out} via an oracle to isolate representation.
\textbf{Hierarchical retrieval.} RAPTOR \citep{raptor} and GraphRAG \citep{graphrag} build
summary hierarchies; our probe results bound what summary nodes can carry for \emph{acting}, and
in exploratory tests architecture-grounded descent did not beat flat retrieval
(Appendix~\ref{app:explore}).
\textbf{The bitter lesson for code context.} ContextBench \citep{contextbench} reports
sophisticated scaffolds barely beating simple baselines, with much explored context going
unutilized; our results corroborate this from an independently built, audited instrument.
CodeCompass \citep{codecompass} argues the opposite for \emph{navigation} --- structural
traversal beating retrieval when lexical overlap is low --- consistent with our scope: our nulls
concern \emph{act}-context rendering, not navigation.

\section{Method}
\label{sec:method}

\subsection{Experimental setup}
\label{sec:setup}

\begin{table}[h]
  \centering\small
  \caption{Setup. Every claim in the paper inherits these conditions unless stated.}
  \label{tab:setup}
  \begin{tabular}{@{}p{0.26\linewidth}p{0.68\linewidth}@{}}
    \toprule
    Resolve dataset & SWE-bench Verified \citep{swebench}: \textbf{all 71 multi-file instances}
    ($\geq 2$ gold files); 70 after gold-validation exclusion (\texttt{astropy-8707}, whose gold
    patch fails its own environment). \\
    Probe data & 15 held-out classes from seaborn/pylint/pytest at pinned commits, each with a
    dedicated test file (probes minted from tests, never from the scored text). \\
    Agent (resolve) & Claude Sonnet 4.6, single-shot, temperature 0, \texttt{max\_tokens}=8000,
    $\leq 2$ apply-failure retry rounds; reasoning permitted before edit blocks. \\
    Probe judge & GPT-4o-mini (independent family), Claude as second judge
    ($\kappa$ reported). \\
    Patches & Deterministic SEARCH/REPLACE $\to$ \texttt{difflib} unified diff; ambiguous or
    non-matching edits fed back for $\leq 2$ retries; empty SEARCH creates new files. \\
    Resolve & SWE-bench 4.1.0 Docker harness (FAIL\_TO\_PASS $\wedge$ PASS\_TO\_PASS).
    \textbf{Gold patch validated for every instance} (70/71 pass). \\
    Expressibility gate & Per arm$\times$instance, every gold hunk's deleted lines (or insertion
    anchor) verified verbatim-present in the rendered context before any agent call: 70/70 in all
    arms. \\
    Cost & \texttt{cl100k\_base} token proxy, plus Anthropic \texttt{count\_tokens} on the exact
    \emph{initial} prompt per episode (retry prompts are logged in metrics but not token-counted). \\
    Statistics & Wilson 95\% CIs on every proportion; exact two-sided McNemar on paired resolve. \\
    \bottomrule
  \end{tabular}
\end{table}

\textbf{Structural tree.} Deterministic \texttt{ast} extraction yields
Subsystem$\to$Component$\to$File$\to$Class$\to$Function. \textbf{Representation rungs:} full-body
$\cdot$ method-only $\cdot$ UML-class-skeleton $\cdot$ signature+doc $\cdot$ name $\cdot$ exclude
(all deterministic; no LLM).

\subsection{The \resolveatcost{} rig}
\label{sec:rig}

\begin{figure}[t]
  \centering
  \includegraphics[width=\linewidth]{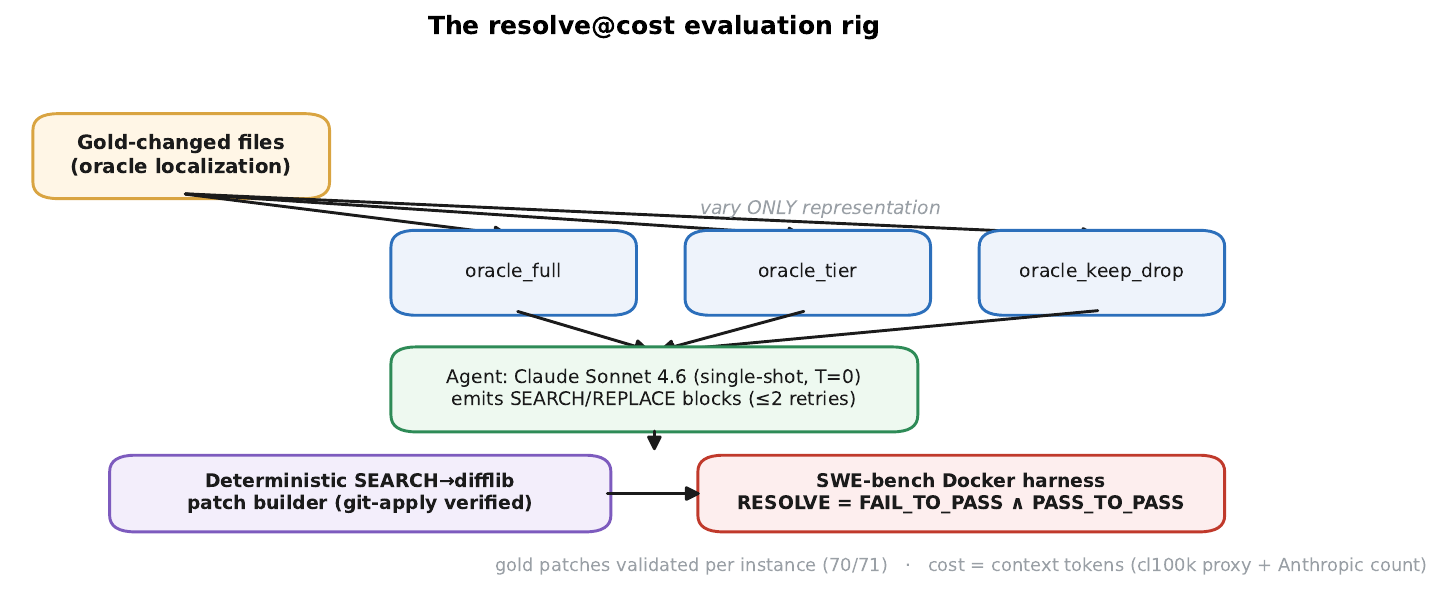}
  \caption{\textbf{The rig, end to end.} Gold-changed files (oracle localization) are rendered
  under three arms that vary representation; the agent emits SEARCH/REPLACE blocks; a
  deterministic builder constructs the diff; the official Docker harness reports resolution.}
  \label{fig:rig}
\end{figure}

\textbf{Oracle localization} isolates representation from retrieval: arms share the gold-changed
files. The arms are also \emph{coverage-fair}: the compressed arms render module-level
code, class-attribute regions, decorators, and non-Python gold files verbatim, so every gold edit
is expressible from every arm's context (verified per instance; Table~\ref{tab:setup}).
We disclose the asymmetry: the compressed arms are built from gold-edited units (they encode
\emph{where} to edit at full detail); \texttt{oracle\_full} receives whole files with no
highlighting. All arms are oracle \emph{ceilings}, not deployable policies; the
tier-vs-keep/drop comparison --- which shares the same gold units --- is the clean one.

\textbf{Deterministic patches} remove the main confound of LLM-authored diffs: in development
testing, model-authored unified diffs failed to apply on 7 of the 9 attempts; the
SEARCH$\to$\texttt{difflib} builder applies cleanly, \texttt{git apply --check}-verified, with
uniqueness checking, new-file creation, and $\leq 2$ retry rounds with per-edit failure feedback.

\textbf{\resolveatcost{}} is reported as the pair (resolve rate, mean context tokens) and
summarized as \emph{tokens per resolve} $=$ total context tokens / instances resolved. We do not
divide resolve by mean tokens: that ratio rewards an arm for shrinking context on instances it
never solves.

\section{The find/act boundary}
\label{sec:findact}

Can a compressed stand-in for source code --- a summary, a signature --- carry what an agent needs
to \emph{act}? We measure this with behavioral probes: 15 held-out classes from
\textbf{seaborn / pylint / pytest} at pinned commits, each with a dedicated test file; three probes
per class minted \emph{from the tests} (independent behavioral ground truth --- the answerer never
sees them; ``UNKNOWN'' permitted); a single frontier model (Claude Sonnet 4.6) reads the
representation and answers every probe in every arm; answers graded by an \emph{independent-family}
judge (GPT-4o-mini), with a same-family judge run in parallel only to report agreement
($\kappa{=}0.84$, 94\% raw). Temperature 0 throughout. A summarizer-quality control runs the same
prompt at the same 180-token budget with both a small model (Qwen2.5-Coder-3B) and a frontier
model (Claude Sonnet 4.6), so ``summaries fail'' can't be an artifact of a weak summarizer.

\begin{table}[t]
  \centering\small
  \caption{\textbf{The find/act boundary} ($n{=}45 = 15$ held-out classes $\times$ 3 test-minted
  probes; independent judge; strict $T{=}0$). Exact McNemar vs.\ source: 23-vs-0 discordant for
  both summary arms, 21-vs-0 for signature (all $p < 10^{-4}$).}
  \label{tab:findact}
  \begin{tabular}{@{}lccr@{}}
    \toprule
    context & accuracy & Wilson 95\% CI & tokens \\
    \midrule
    full source & \textbf{0.600} (27/45) & \ci{0.45}{0.73} & 1{,}124 \\
    summary --- frontier (Sonnet 4.6) & 0.089 (4/45) & \ci{0.04}{0.21} & 165 \\
    summary --- Qwen-3B & 0.089 (4/45) & \ci{0.04}{0.21} & 180 \\
    signature + docstring & 0.133 (6/45) & \ci{0.06}{0.26} & 212 \\
    \bottomrule
  \end{tabular}
\end{table}

Three observations. First, \textbf{source is not perfect} ($0.600$): test-minted behavioral probes
are hard, and sometimes need context beyond one class --- which is exactly why a fair protocol
matters: a probe set minted from the scored text itself reports a flattering $1.00$
(Appendix~\ref{app:explore}). Second, \textbf{the boundary is ${\sim}7\times$ wide}: source answers $60\%$ of the questions,
summaries answer only ${\sim}9\%$, and no summary ever succeeded where source failed. Third ---
the control that gives the result its teeth --- \textbf{frontier-written summaries score
identically to 3B summaries}: the gap is a property of the \emph{representation class}, not
summarizer quality. Better prose does not buy actionable information at this budget. Signatures,
at similar cost, edge out prose summaries ($6/45$ vs.\ $4/45$) --- at these counts no more than a hint that what
little survives compression is structure, not narrative.

This says summaries cannot \emph{act}; it does \textbf{not} establish that summaries help
\emph{navigate} --- in our exploratory navigation test (Appendix~\ref{app:explore}), summary-keyed
hierarchical descent did not beat flat retrieval either.

\section{Resolve@cost: every multi-file Verified instance, pre-registered}
\label{sec:resolve}

\textbf{Protocol (frozen and committed before any data).} All 71 instances whose gold patch
touches $\geq 2$ files; oracle localization; vary only representation; single-shot agent
(Table~\ref{tab:setup}). Every instance's gold patch is first run through the harness ($70/71$
environments pass; the failure is excluded and recorded), and every gold edit is verified
expressible in every arm before any agent call. Registered hypothesis H2:
\texttt{oracle\_tier} resolves more than \texttt{oracle\_keep\_drop}.

\begin{figure}[t]
  \centering
  \includegraphics[width=0.9\linewidth]{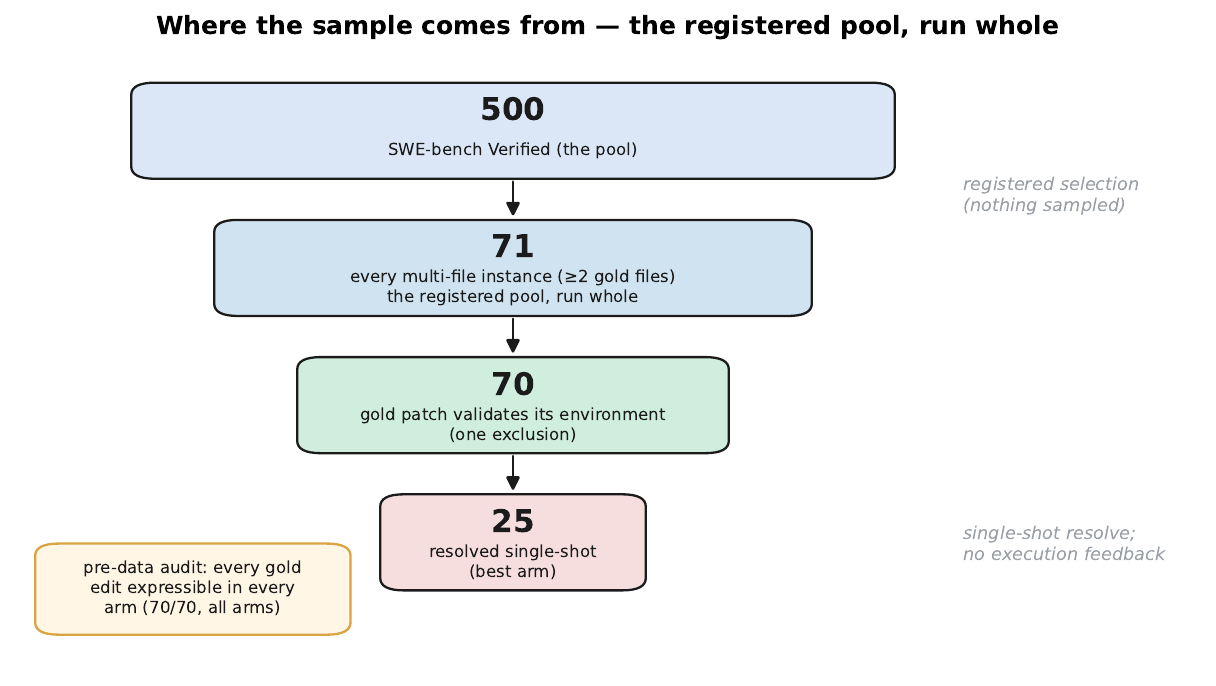}
  \caption{\textbf{Where the sample comes from.} 500 Verified instances $\to$ 71 multi-file (the
  registered criterion) $\to$ 70 after one gold-validation exclusion. The whole pool runs; nothing
  is sampled.}
  \label{fig:selection}
\end{figure}

\begin{figure}[t]
  \centering
  \includegraphics[width=\linewidth]{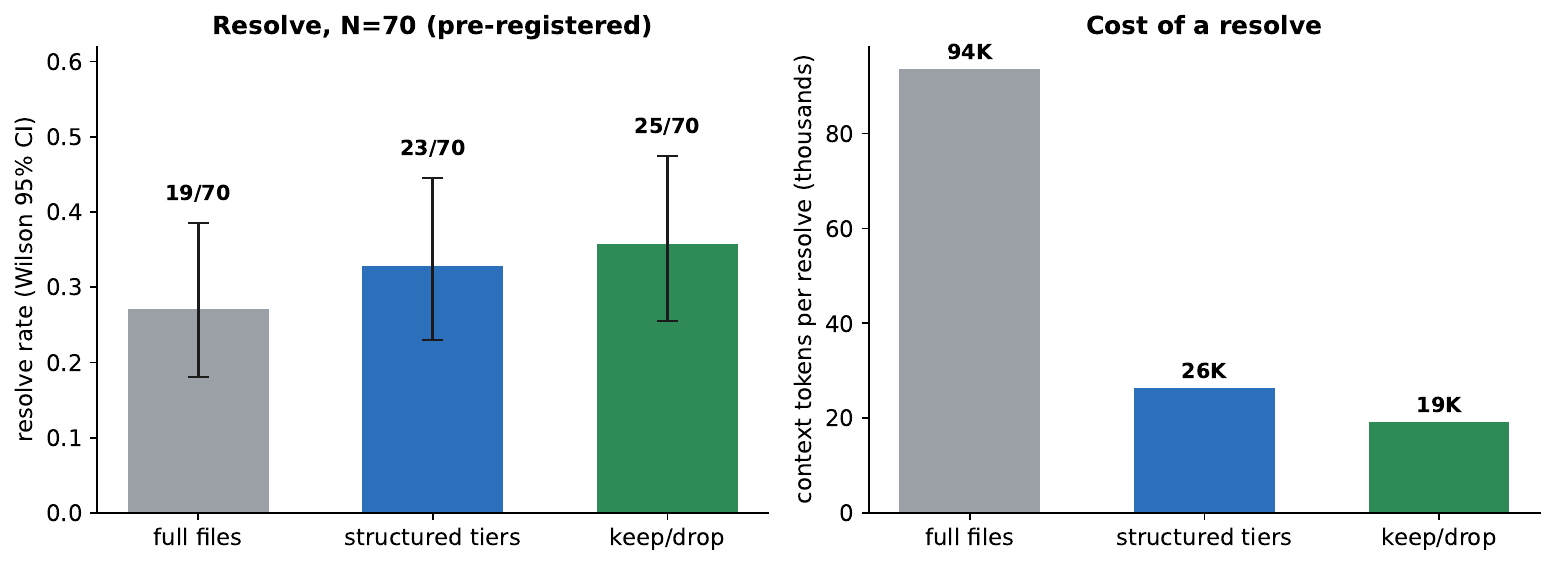}
  \caption{\textbf{The powered result ($N{=}70$).} \emph{Left:} resolve rate per arm with Wilson
  95\% CIs --- the registered tier-vs-keep/drop comparison is null (exact McNemar $p{=}0.754$).
  \emph{Right:} context tokens per resolved instance: compressed context buys a resolve for
  $19$--$26$K tokens; whole files pay $94$K.}
  \label{fig:powered}
\end{figure}

\begin{table}[t]
  \centering\small
  \caption{Resolve and context cost ($N{=}70$; every arm covers every gold hunk on every instance,
  so differences are due to representation, not expressibility).}
  \label{tab:resolve}
  \begin{tabular}{@{}lcccr@{}}
    \toprule
    arm & resolved & rate \ci{\text{95\% CI}}{} & mean ctx tokens & tokens/resolve \\
    \midrule
    \texttt{oracle\_full} & 19/70 & 0.271 \ci{0.18}{0.39} & 25{,}426 & 93{,}700 \\
    \texttt{oracle\_tier} & 23/70 & 0.329 \ci{0.23}{0.44} & 8{,}629 & 26{,}300 \\
    \texttt{oracle\_keep\_drop} & \textbf{25/70} & \textbf{0.357} \ci{0.26}{0.47} & \textbf{6{,}876} & \textbf{19{,}300} \\
    \bottomrule
  \end{tabular}
\end{table}

\begin{table}[t]
  \centering\small
  \caption{Paired tests (exact McNemar, two-sided). The registered H2 is null.}
  \label{tab:mcnemar}
  \begin{tabular}{@{}lccc@{}}
    \toprule
    comparison & A-only & B-only & $p$ \\
    \midrule
    tier vs.\ keep/drop (\textbf{H2, registered}) & 4 & 6 & \textbf{0.754} \\
    tier vs.\ full & 7 & 3 & 0.344 \\
    keep/drop vs.\ full & 7 & 1 & 0.070 \\
    \bottomrule
  \end{tabular}
\end{table}

\paragraph{Findings.}
\textbf{(1) The registered hypothesis is null: structured tiers do not beat binary keep/drop}
($p{=}0.75$; keep/drop is nominally \emph{ahead}). Rendering the non-locus remainder as class
skeletons and signatures carries no measurable resolve signal over just dropping it, at oracle
localization, single-shot. \emph{Scope:} the tiers tested are the thin end of the
structured-artifact family --- static name/signature digests; interaction-level artifacts and
distance-graded detail were specified in our original design but never implemented (they require
call-edge extraction this codebase lacks); the null does not cover them.
\textbf{(2) Compressed context resolves at least as well as whole files at $3$--$3.7\times$ fewer
tokens.} Both compressed arms are nominally above full; keep/drop vs.\ full is marginal
($p{=}0.07$) and we do not claim significance. On cost-normalized terms the gap is clear:
$19{,}300$ context tokens per resolve vs.\ $93{,}700$ (Figure~\ref{fig:powered}).
\textbf{(3) Format failures do not drive outcomes:} 0--1 truncated generations and 1--2 empty
patches per arm; the retry loop fired on 11--24 instances per arm (most often for tier, whose
synthesized renderings tempt the model to copy non-verbatim text --- a residual asymmetry we
disclose; it cuts against H2, which was null anyway).

\paragraph{Interaction-level follow-up (pre-registered).} Since the tiers above are static
digests, we separately tested the sharpest \emph{interaction} artifact: keep/drop's byte-identical
content plus a deterministic AST-derived sequence slice of the gold functions' local call flow
(present on 65/70 instances, mean 215 tokens). Registered H3: slice $>$ keep/drop. Result:
\textbf{directionally positive, not significant} --- $26/70$ ($0.371$ \ci{0.27}{0.49}) vs.\
$23/70$ ($0.329$ \ci{0.23}{0.44}), McNemar 4-vs-1, $p{=}0.375$. The only structured artifact in
the study that pointed the right way; at $N{=}70$ it supports no claim.

\paragraph{Run-to-run noise bounds all small gaps.} The follow-up re-ran the keep/drop arm under
the byte-identical protocol (temperature 0, same model alias, same prompts): it resolved $23/70$
vs.\ the original $25/70$, with a symmetric difference of \textbf{6 instances}. Temperature-0 API
inference is not deterministic; ${\sim}9\%$ of per-instance outcomes flip between identical runs.
Any arm difference of this order --- including H2's 2-instance gap --- is within run noise, a
caution that applies to single-run SWE-bench comparisons generally.

\section{Threats to validity (and which way each cuts)}
\label{sec:threats}

\textbf{Single-shot, no-execution regime.} In this regime there is one completion (plus
apply-failure retries) --- no reproduction script, no test run, no repair loop. This is the regime \emph{least} likely to
surface representation effects --- a multi-turn agent can recover from compressed context by
reading more --- so the H2 null is a statement about this regime; it biases toward the null for
tier-vs-keep/drop and plausibly \emph{against} full. Conclusions about \emph{agents} require the
in-the-loop version (future work; Appendix~\ref{app:explore}).
\textbf{Oracle ceilings, asymmetric construction.} Compressed arms encode gold edit
\emph{locations}; full does not. Tier-vs-keep/drop shares the same gold units and is clean;
compressed-vs-full mixes compression with location hinting --- one reason we don't claim the
$p{=}0.07$ comparison. Realistic retrieval would lower every arm together.
\textbf{Probe scope.} The find/act boundary rests on held-out repos, test-minted probes, and an
independent judge (\S\ref{sec:findact}), but on one language and 15 classes. Because probes
cluster within classes, we also computed class-clustered bootstrap intervals: source
$\ci{0.40}{0.78}$ vs.\ summaries $\ci{0.00}{0.20}$, gap CI $\ci{0.36}{0.67}$ --- the boundary is
unchanged under clustering; the exploratory
own-code probes in Appendix~\ref{app:explore} show how much a circular protocol inflates positive
controls ($1.00 \to 0.600$ for source) and are not relied on for any claim.
\textbf{Format asymmetry, residual.} Tier's synthesized text triggered the most edit retries
($24/70$ vs.\ $15$ keep/drop, $11$ full); gold-edit regions are always verbatim, but a residual
format penalty against tier can't be fully excluded (it cuts against H2, which was null).
\textbf{Non-determinism --- measured, not assumed.} $T{=}0$ flips ${\sim}9\%$ of outcomes between
identical runs (\S\ref{sec:resolve}); probe cells are single draws at $T{=}1$.
\textbf{Token units.} Headline counts are a GPT-BPE proxy; Anthropic counts are logged alongside
(mean full prompt: full $32.6$K / tier $11.6$K / keep/drop $9.4$K). Ratios are robust; absolutes
differ.
\textbf{Single language (Python), single model family.} Stated coverage gaps.
\textbf{Provenance.} Protocols, hypotheses, and samples were committed before data collection;
figures regenerate from raw result files; per-instance harness reports ship with the artifact.

\section{Conclusion}
\label{sec:conclusion}

On every multi-file instance in SWE-bench Verified, with localization fixed and every gold edit
verified expressible in every arm, the answer to ``what context does a single-shot agent need to
\emph{act}'' is: \textbf{the edited code itself --- and neither of the surrounding-context
renderings we tested (class skeletons, signatures) adds anything over dropping it.} Our registered
hypothesis, that structured tiers preserve resolve signal that binary keep/drop discards, is null
at $N{=}70$, and both compressed arms resolve at least as well as whole files at $3$--$3.7\times$
fewer tokens. The same was true of every other measurement we ran --- descent vs.\ flat,
ownership scoring, summaries vs.\ source: each structure-aware elaboration failed to beat the
simpler alternative. We see this as one more controlled data point for the bitter lesson in code
context, with the usual scope limits (single-shot, oracle ceilings, one model family, Python). The
lasting contributions are the audited \resolveatcost{} rig and the discipline of registering
directional hypotheses and publishing their nulls.

\bibliographystyle{plainnat}
\bibliography{refs}

\appendix

\section{Agent prompt and decoding}
\label{app:prompt}

The agent receives, for every arm, the byte-identical instruction below; only \texttt{\{context\}}
varies. Decoding (v2): \texttt{max\_tokens}=8000, temperature 0, no truncation of issue or
context; $\leq 2$ retry rounds add the failed edits' diagnostics (not-found / ambiguous /
no-such-file) and ask for corrected blocks only.

\begin{small}
\begin{verbatim}
You are fixing a GitHub issue in the `{repo}` repository.
ISSUE: {issue}
RELEVANT CODE (already localized for you): {context}
Think briefly about the fix first if that helps. Then output ONE OR MORE
edit blocks in EXACTLY this format (the blocks are parsed mechanically;
everything outside them is ignored):
*** FILE: <path relative to repo root>
*** SEARCH
<exact existing lines to replace -- copy them VERBATIM from the code
above, incl. indentation>
*** REPLACE
<the new lines>
*** END
Rules:
- each SEARCH block must be copied verbatim from the provided code and
  match exactly ONE location; keep it small and uniquely matching.
- to CREATE a new file, emit a block whose SEARCH section is empty.
- edit only what is necessary to fix the issue.
\end{verbatim}
\end{small}

\section{Reproducibility, resources, and research transparency}
\label{app:repro}

\paragraph{Artifact.} All experiment code, frozen instance lists, pre-registrations, per-instance
harness reports, and the figure generator (every plotted number loaded from a raw result file) are
released at \url{https://github.com/integrallis/act-context} under the MIT license.
The protocol, frozen sample, and expressibility audit were committed before any agent call. Known
limits: the model id resolves to an alias (\texttt{claude-sonnet-4-6}; the API exposes no dated
snapshot or seed); probe cells are single samples; headline token counts are a
\texttt{cl100k\_base} proxy (Anthropic counts cover each episode's initial prompt; retry prompts
are logged but not token-counted). One operational note: the first
evaluation pass hit Docker Hub's anonymous image-pull rate limit; affected instances were detected
as \emph{environment} errors (never counted as model failures) and re-run to completion under
authenticated pulls.

\paragraph{Compute and cost.} The resolve experiments are API calls plus commodity Docker: the
registered $N{=}70$ run used ${\approx}210$ agent calls (${\approx}\$20$ of API inference) and
${\approx}6$ hours of Docker evaluation wall-clock on a 32-thread workstation; the sequence-slice
follow-up roughly doubles the evaluation half; the probe study (\S\ref{sec:findact}) used
${\approx}600$ short model calls (${\approx}\$5$). No GPUs and no training are required anywhere;
figures regenerate on CPU in seconds.

\paragraph{Existing assets.} SWE-bench Verified and its evaluation harness \citep{swebench} (MIT
license); the held-out probe corpora are public open-source repositories (seaborn, pylint,
pytest), analyzed at pinned commits under their respective licenses. Models: Anthropic Claude
Sonnet 4.6 (agent, summarizer control, second judge), OpenAI GPT-4o-mini (independent judge), and
Qwen2.5-Coder-3B (Apache-2.0; small-summarizer arm), each used via its terms of service.

\paragraph{LLM usage declaration.} LLMs are the \emph{subject} of this paper: the evaluated agent,
the summarizer arms, and the judges are all LLMs, fully specified in Table~\ref{tab:setup} and
\S\ref{sec:findact}. In addition, LLM-based development tooling assisted in implementing the
experimental infrastructure and drafting the manuscript; all hypotheses, protocols, and analysis
decisions were human-approved and frozen in committed pre-registrations before data collection,
and every reported number is a deterministic artifact of the released code and raw result files.

\paragraph{Broader impact.} This is foundational measurement on a public benchmark. Its main
practical implication --- that agents resolve issues equally well with a third of the context ---
points toward cheaper, lower-energy coding agents; we see no direct path to misuse, and the
released assets (evaluation code and result files) pose no dual-use risk.

\section{Pre-registrations and their outcomes}
\label{app:prereg}

\textbf{Scaled resolve (executed).} Frozen before any data: all 71 multi-file instances;
coverage-fair arms; gold validation per instance; expressibility instrumentation; Wilson +
exact McNemar; H2 (tier $>$ keep/drop) registered, a tie reported as the result. Outcome:
\S\ref{sec:resolve} --- H2 null ($p{=}0.754$).
\textbf{Probe re-grade (executed).} Held-out classes with dedicated test files; probes minted from
tests; independent-family judge with $\kappa$; frontier-summarizer control; $T{=}0$. Outcome:
\S\ref{sec:findact} --- the boundary hardens (source $27/45$ vs.\ summaries $4/45$, both
summarizers identical; canonical numbers from the corrected strict-$T{=}0$ rerun).
\textbf{Sequence slice (executed).} Keep/drop content $+$ AST call slice, both arms fresh;
H3 (slice $>$ keep/drop). Outcome: directionally positive, ns ($p{=}0.375$); the keep/drop re-run
doubles as the run-to-run noise measurement (6/70 flips).
\textbf{Registered escalation (future):} a 2-step agent loop on the solvable subset; an act-view
re-grade under the \S\ref{sec:findact} protocol.

\section{Exploratory results (not claims of this paper)}
\label{app:explore}

This appendix collects superseded measurements that ran on the project's \emph{own in-house
codebases} --- small, self-authored corpora --- and/or under self-judged protocols. They are
reported for transparency and instrument history; no claim in the main text relies on them.

\paragraph{(a) Pilot probe protocol (superseded by \S\ref{sec:findact}).} Probes minted from the
scored source (``answerable only from this source''), answered and graded by the same model, on
in-house code ($n{=}15$): source $15/15$, summaries $0/15$. Under the honest protocol of
\S\ref{sec:findact}, the source control falls to $0.600$ --- a measure of how much circular
minting and self-judging inflate positive controls --- while the boundary itself survives.

\paragraph{(b) Act-view ladder (own code, self-judged, construction-guaranteed positives).}
Which rendering suffices to act on a single method ($n{=}18 = 6$ methods $\times$ 3 probes)?
Full class $1.00$ at $18{,}661$ tokens; UML-skeleton$+$body $1.00$ at $2{,}814$; \textbf{method
body alone $1.00$ at $900$} (${\approx}20\times$ fewer than the class); signature-only $0.06$.
Suggestive that the body is the budget --- but \texttt{method\_only}$=1.00$ is partly guaranteed
by construction (the answerer sees the source the probe was minted from), so we treat this as a
hypothesis for a future test-minted replication, not a finding.

\paragraph{(c) Hierarchical descent vs.\ flat retrieval (own code, $n{=}16$ queries).}
Summary-keyed descent over an architecture tree did not beat flat retrieval (flat $0.875$
\ci{0.64}{0.97} vs.\ cumulative-summary descent $0.812$ \ci{0.57}{0.93}); residual-summary keys
were worse ($0.562$). Untested at the repository scales where hierarchy could matter.

\end{document}